\documentclass[conference]{IEEEtran}
\IEEEoverridecommandlockouts

\usepackage{cite}
\usepackage{amsmath,amssymb,amsfonts}
\usepackage{algorithmic}
\usepackage{graphicx}
\usepackage{textcomp}
\usepackage{xcolor}

\usepackage{CJKutf8}
\usepackage{subcaption}

\usepackage{balance} 

\usepackage{multirow}
\usepackage{makecell}
\usepackage{array} 

\def\BibTeX{{\rm B\kern-.05em{\sc i\kern-.025em b}\kern-.08em
    T\kern-.1667em\lower.7ex\hbox{E}\kern-.125emX}}
\begin{document}

\begin{CJK}{UTF8}{gbsn}

\title{Agentic Copyright Watermarking against\\Adversarial Evidence Forgery with Purification-Agnostic Curriculum Proxy Learning
\thanks{This work was partially supported by JSPS KAKENHI Grants JP21H04907 and JP24H00732, by JST CREST Grants JPMJCR18A6 and JPMJCR20D3 including AIP challenge program, by JST AIP Acceleration Grant JPMJCR24U3, and by JST K Program Grant JPMJKP24C2 Japan.} 
}

\makeatletter
\newcommand{\linebreakand}{
  \end{@IEEEauthorhalign}
  \hfill\mbox{}\par
  \mbox{}\hfill\begin{@IEEEauthorhalign}
}
\makeatother

\author{\IEEEauthorblockN{Erjin Bao}
\IEEEauthorblockA{\textit{Graduate University for Advanced Studies, SOKENDAI} \\
\textit{National Institute of Informatics}\\
Tokyo, Japan \\
bao-erjin@nii.ac.jp}
\and
\IEEEauthorblockN{Ching-Chun Chang}
\IEEEauthorblockA{\textit{Information and Society Research Division} \\
\textit{National Institute of Informatics}\\
Tokyo, Japan \\
ccchang@nii.ac.jp}
\linebreakand
\IEEEauthorblockN{Hanrui Wang}
\IEEEauthorblockA{\textit{Information and Society Research Division} \\
\textit{National Institute of Informatics}\\
Tokyo, Japan \\
hanrui\_wang@nii.ac.jp}
\and
\IEEEauthorblockN{Isao Echizen}
\IEEEauthorblockA{\textit{Information and Society Research Division} \\
\textit{National Institute of Informatics}\\
Tokyo, Japan \\
iechizen@nii.ac.jp}
}

\maketitle

\begin{abstract}
With the proliferation of AI agents in various domains, protecting the ownership of AI models has become crucial due to the significant investment in their development. Unauthorized use and illegal distribution of these models pose serious threats to intellectual property, necessitating effective copyright protection measures. Model watermarking has emerged as a key technique to address this issue, embedding ownership information within models to assert rightful ownership during copyright disputes. This paper presents several contributions to model watermarking: a self-authenticating black-box watermarking protocol using hash techniques, a study on evidence forgery attacks using adversarial perturbations, a proposed defense involving a purification step to counter adversarial attacks, and a purification-agnostic curriculum proxy learning method to enhance watermark robustness and model performance. Experimental results demonstrate the effectiveness of these approaches in improving the security, reliability, and performance of watermarked models.
\end{abstract}

\begin{IEEEkeywords}
adversarial evidence forgery, agentic copyright, model watermarking, proxy learning.
\end{IEEEkeywords}

\section{Introduction}
With advances in neural network technology, AI agents are now widely used in various fields, including computer vision~\cite{Girshick2015,Redmon2016,Ronneberger2015}, speech recognition~\cite{Hinton2012}, and language generation~\cite{Vaswani2017,Radford2019,Devlin2019}. These models are embedded in applications and websites, and their development requires significant investment in data and computation. As valuable intellectual property, protecting these models is essential to prevent unauthorized use and distribution. For example, companies might deploy others' models without permission, or individuals may distribute them illegally for profit. Model watermarking technology addresses these issues by embedding ownership information within the models, enabling clear identification of rightful ownership in copyright disputes~\cite{Uchida2017,Boenisch2021,Li2021}. The focus of recent research has been on two primary watermarking techniques: parameter embedding~\cite{Nagai2018,Kuribayashi2020,Rouhani2019} and trigger set embedding~\cite{Zhang2018,Adi2018,Leroux2024}. Parameter embedding involves inserting watermark information directly into the model's parameters, suitable for white-box scenarios where direct access to model parameters is available. However, this white-box method risks exposing sensitive model details. In contrast, trigger set embedding introduces a special evidence set during training that prompts specific outputs for certain inputs, suitable for black-box verification scenarios where the verifier only needs access to the model's outputs. This black-box approach better safeguards model confidentiality and is more versatile in real-world applications. The main contributions are as follows:
\begin{itemize}
    \item \textbf{Established a self-authenticating black-box model watermarking protocol}: This protocol uses hash techniques to enable self-authentication without requiring watermark registration, simplifying the verification process and enhancing system flexibility.

    \item \textbf{Addressed an evidence forgery attack with adversarial purification}: This work demonstrates how adversarial attacks can forge evidence, posing a security threat to black-box watermarking, and addresses counterfeit attacks with integrated purification mechanisms.
    
    \item \textbf{Proposed a purification-agnostic curriculum proxy learning method}: The proposed proxy learning method mitigates the impact of unseen purification on non-adversarial samples, reducing accuracy degradation and enhancing both watermark verification reliability and overall model performance.
\end{itemize}

\begin{figure}[h]
    \centering
    \begin{subfigure}[b]{0.9\linewidth}
        \centering
        \includegraphics[width=\linewidth]{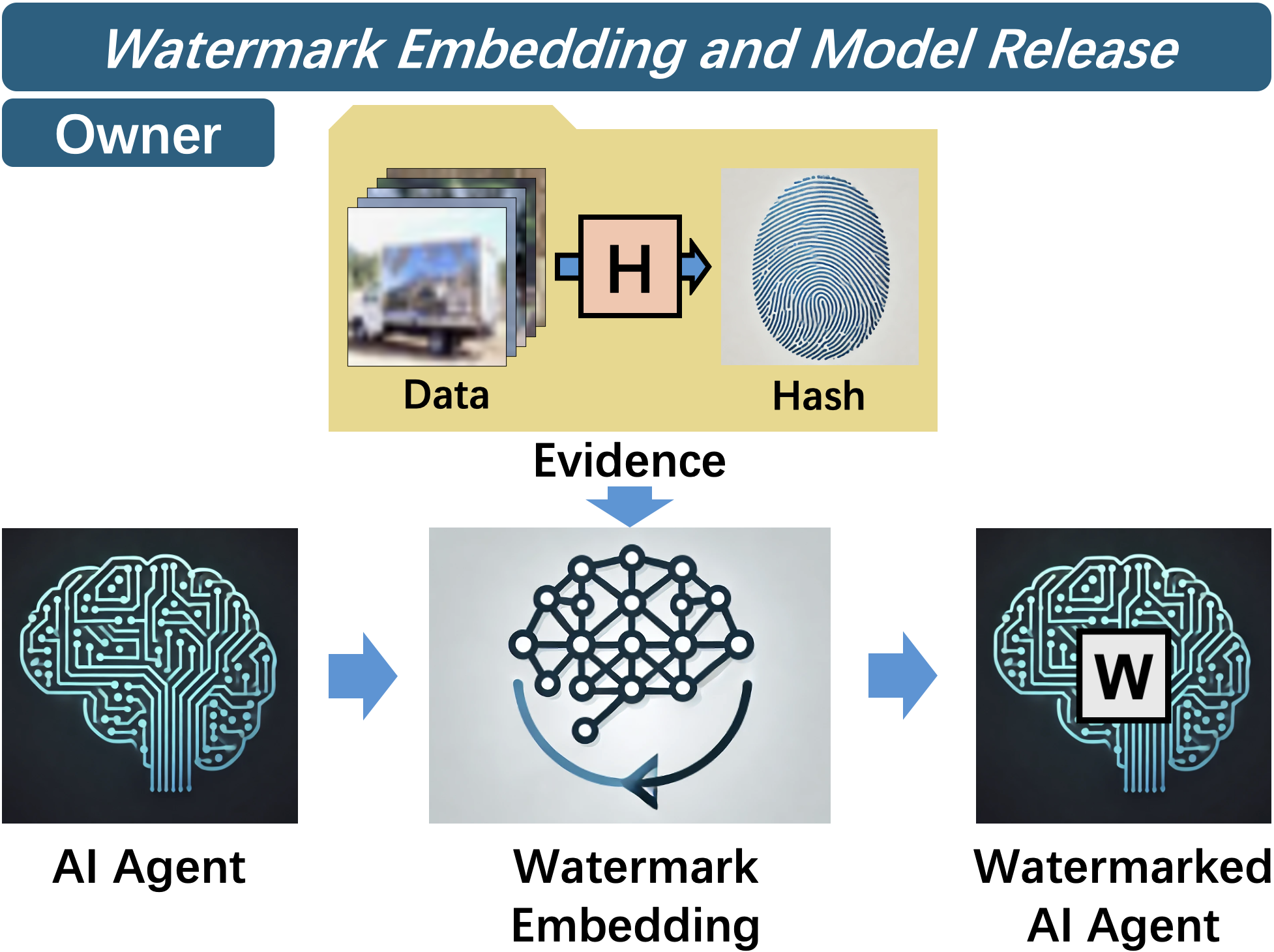}
        \caption{Watermark Embedding}
        \label{fig:pic1}
    \end{subfigure}

    \vspace{0.2cm} 
    
    \begin{subfigure}[b]{0.9\linewidth}
        \centering
        \includegraphics[width=\linewidth]{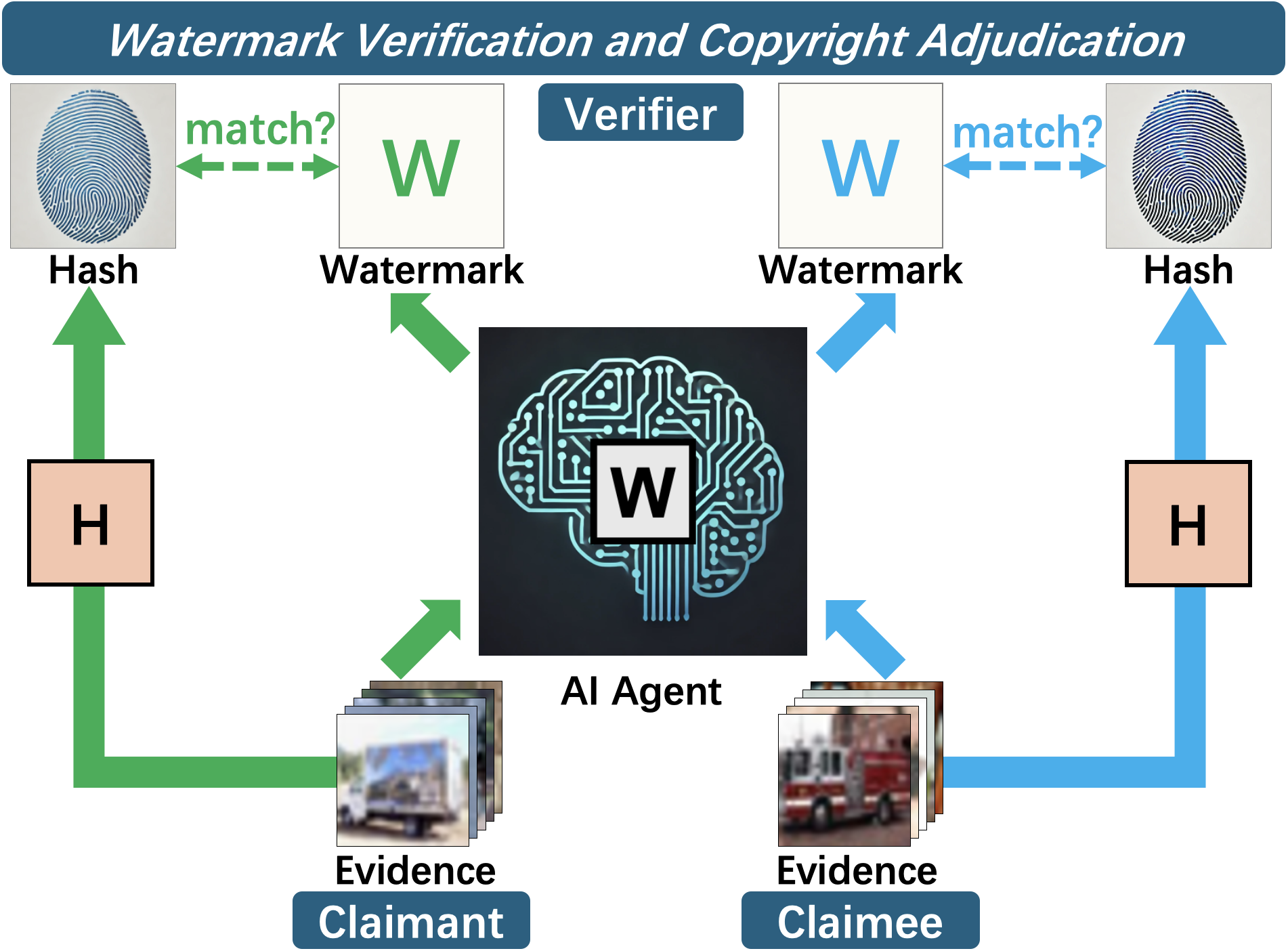}
        \caption{Watermark Verification}
        \label{fig:pic2}
    \end{subfigure}
    \caption{Overview of watermark embedding and verification.}
    \label{fig:combined}
\end{figure}

\section{Methodology}
In this section, we introduce a black-box watermarking protocol and discuss an adversarial evidence forgery and a purification-based countermeasure. Similar to Kerckhoffs principle in cryptography that assumes crypto-algorithms are publicly available while the keys are kept secret, our watermarking protocol follows this principle by assuming our framework and algorithms are transparent to potential attackers while specific parameters used to configure purifiers remain undisclosed. For this reason, we further develop purification-agnostic curriculum proxy learning.

\subsection{Black-Box Watermarking Protocol}
A complete black-box watermarking protocol, consisting of watermark embedding and verification procedures, is shown in Figure~\ref{fig:combined}. The watermark embedding process is described as follows. (\textit{Evidence Standardization}): The owner submits an application to the verifier to obtain a function for creating a standardized evidence as watermark; (\textit{Copyright Protection}): The owner follows the update procedures that fine-tune the model to embed the watermark; (\textit{Model Release}): Once the watermark is embedded, the owner releases the watermarked model for commercial or public use. The watermark verification process is described as follows. (\textit{Copyright Claim}): When there is a suspicion of copyright infringement due to a model's use, a claim can be made to the verifier, along with supporting information to assert ownership; (\textit{Evidence Submission}): The verifier requests both parties involved to provide evidence of ownership; (\textit{Ownership Adjudication}): Based on the submitted evidence, the verifier assesses the watermark to determine the rightful owner and make a final adjudication of ownership. For an image classification model, the function for creating the watermark can be any function that maps an image sample to a class label. The watermark should possess the following properties. First, it should be long enough to avoid collision, which can be achieved by incorporating a sufficient number of image samples into the evidence set. Second, the function should have low probability to map an image to its corresponding class because the watermark is detected by abnormal behaviors/predictions of the model. A hash function can serve as a possible candidate that satisfies the above requirements~\cite{Chi2017}. It can map a set of image samples to a sufficiently long binary sequence, which is then segmented and converted into individual random class labels.

\subsection{Adversarial Evidence Forgery}

\begin{figure}[t!]
    \centering
    \includegraphics[width=1\linewidth]{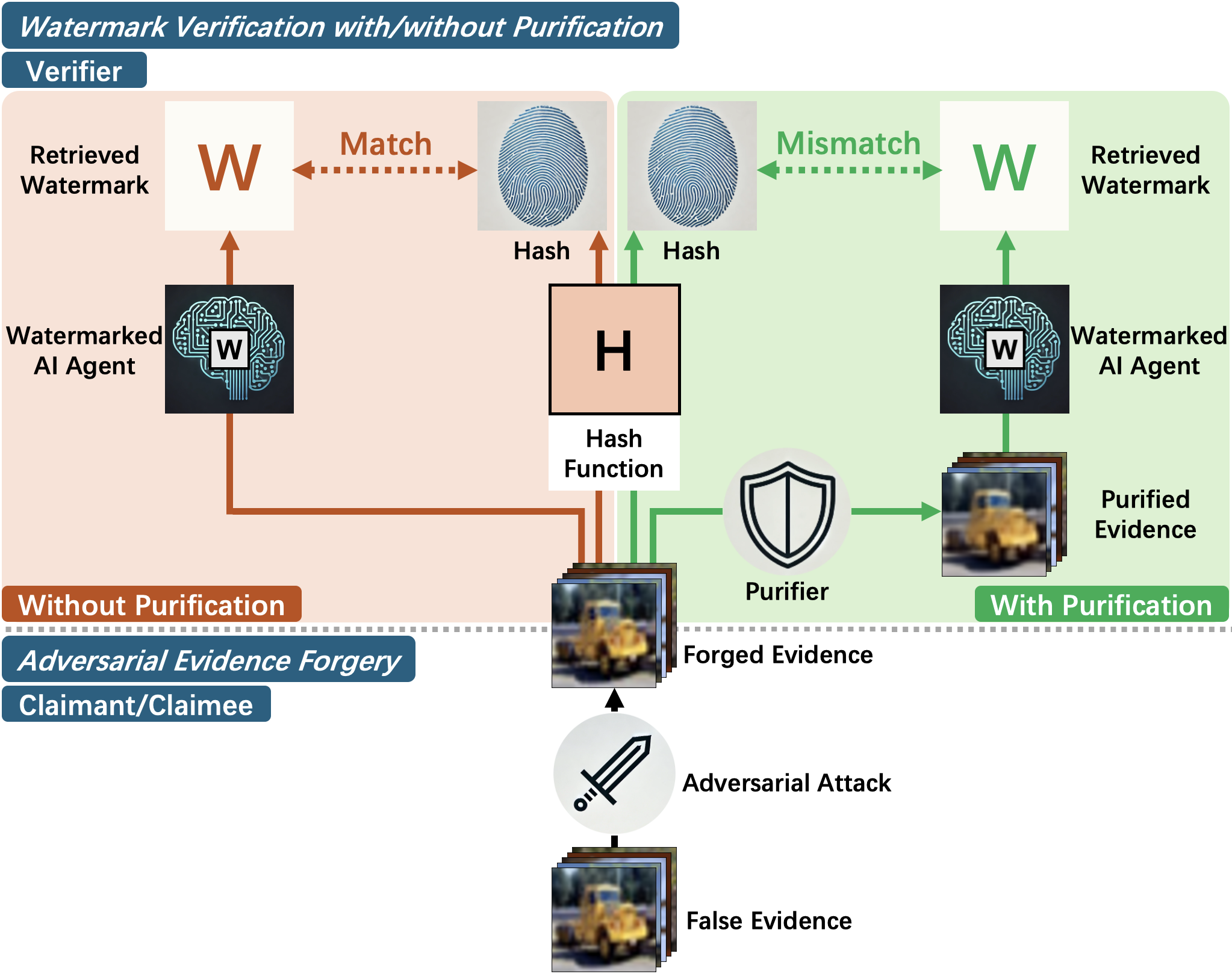}
    \caption{Adversarial evidence forgery and purification.}
    \label{fig:attack_defence}
\end{figure}

Adversarial attacks such as Fast Gradient Signed Method (FGSM)~\cite{Goodfellow2015}, Projected Gradient Descent (PGD)~\cite{Madry2017}, and AutoAttack~\cite{Croce2020} apply small gradient-guided perturbations to input samples to manipulate the model's predictions~\cite{Szegedy2014,Yuan2019,Akhtar2018}. These attack methods can be used to create a set of forged evidence by modifying image samples in a way that their hash values match the classifier's predictions. Additionally, adversarial evidence forgery does not modify the model so that the model's performance is intact. Other attacks such as pruning or re-embedding may require additional data and computational resources and may also cause performance drop because the model is modified. To solve adversarial evidence forgery, we need to incorporate adversarial purification into the verification process as shown in Figure~\ref{fig:attack_defence}. While purification can effectively reduce the impact of adversarial perturbations, it also comes with the side effect of lowering the accuracy in verifying the true evidence~\cite{Guo2018,Liao2018,Samangouei2018,Song2018,Bao2023}. While the hash values are still computed from the original images, the predictions are made from the purified images, which may not be correctly predicted as the pre-defined class labels as purification may also introduce a small amount of distortion to the images.

\begin{figure}[t!]
    \centering
    \begin{subfigure}[b]{1\linewidth}
        \centering
        \includegraphics[width=\linewidth]{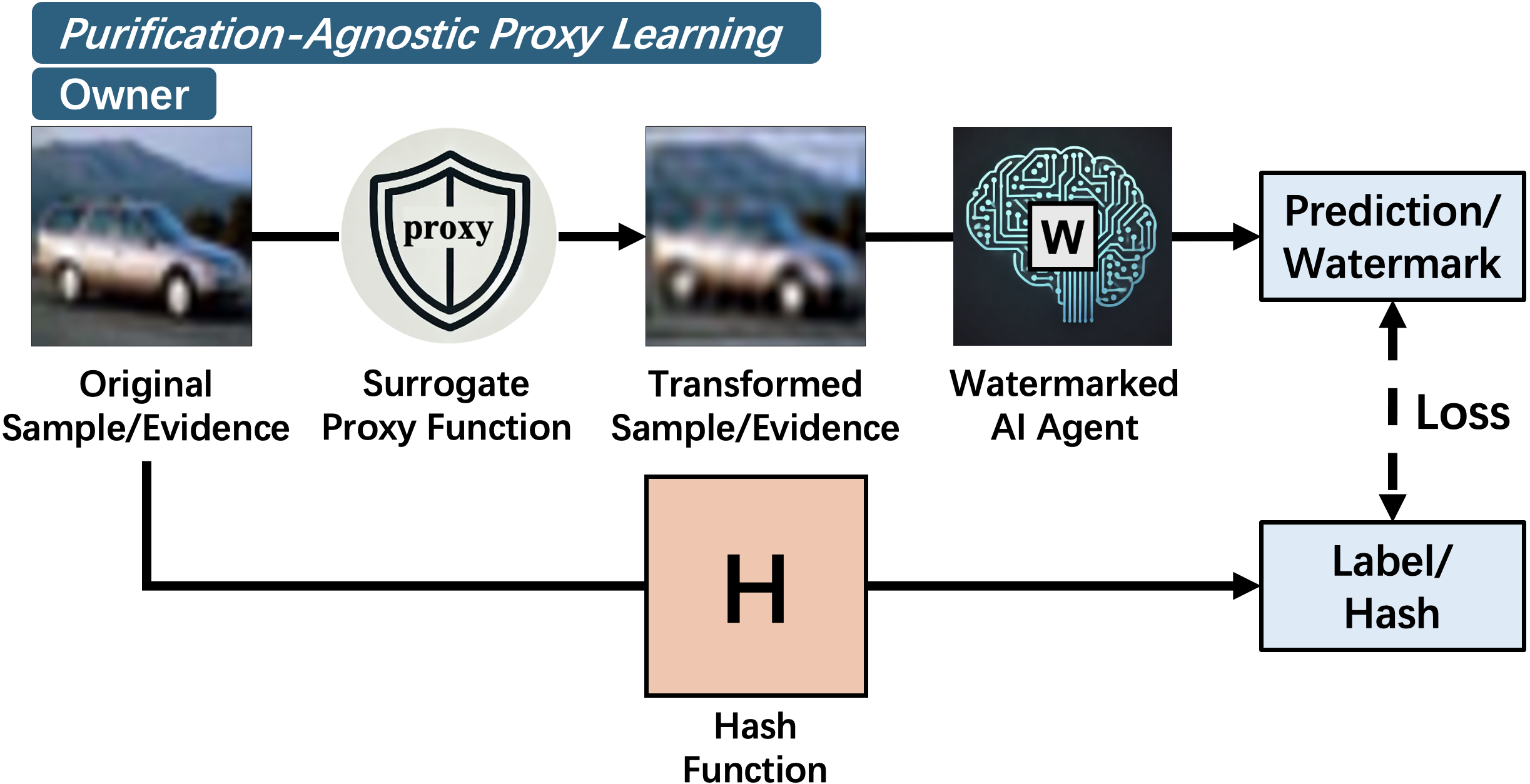}
        \caption{Proxy Learning}
        \label{fig:pic4}
    \end{subfigure}
    
    \begin{subfigure}[b]{0.78\linewidth}
        \centering
        \includegraphics[width=\linewidth]{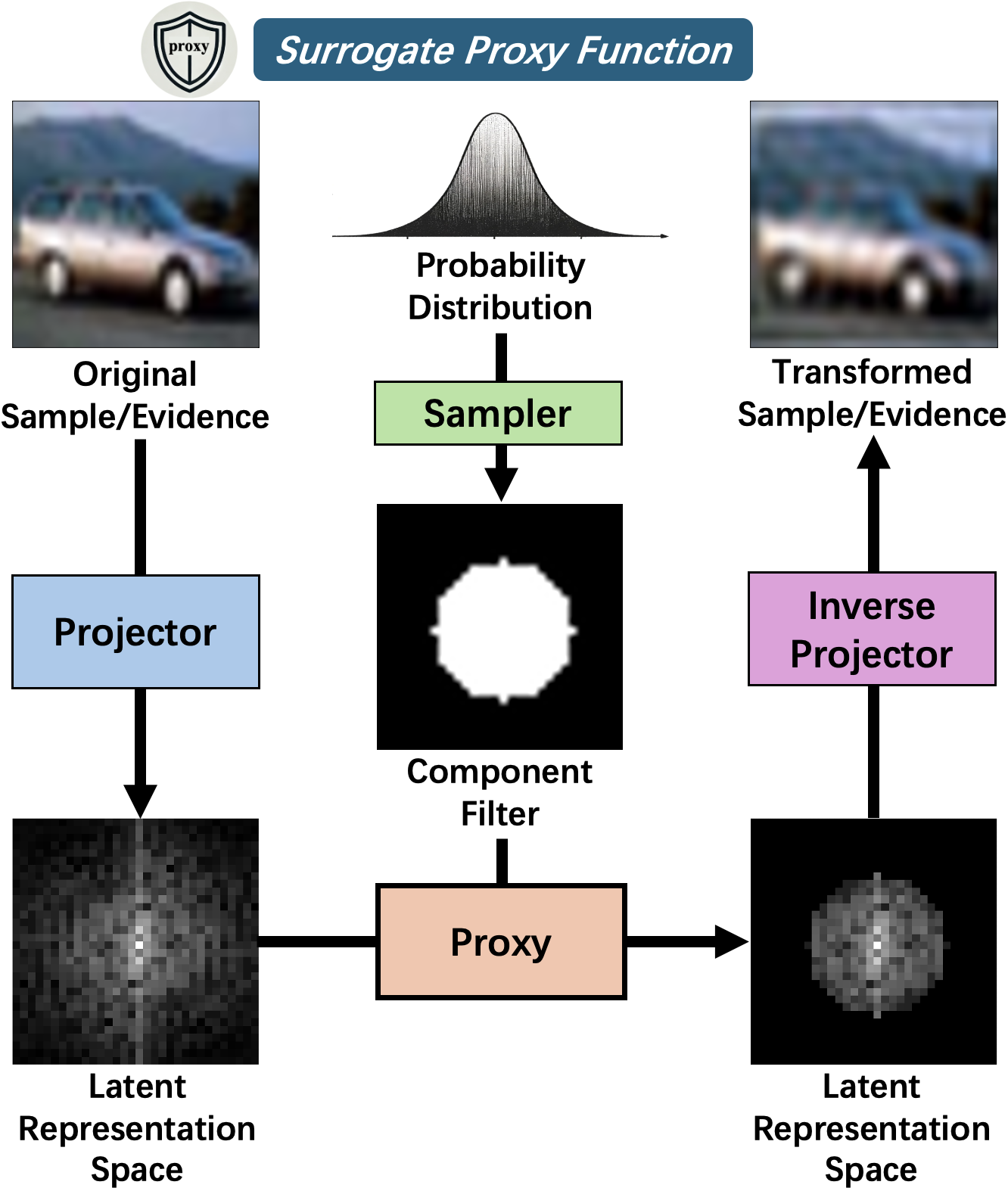}
        \caption{Surrogate Proxy Function}
        \label{fig:pic5}
    \end{subfigure}
    
    \caption{Purification-agnostic curriculum proxy learning.}
    \label{fig:combined2}
\end{figure}

\subsection{Purification-Agnostic Proxy Learning}
The problem of accuracy drop occurs because the distributions of clean images before and after purification are inconsistent, known as distributional shift~\cite{Ganin2015}. Purification is essentially an image transformation process designed to remove harmful adversarial components that impede correct classification, while it may also introduce undesirable distortion to the images. To compensate for this distributional shift, we propose a proxy learning strategy to enhance the model's ability to recognize image content under purification-induced distortion, as illustrated in Figure~\ref{fig:combined2}.
Central to this learning strategy is a surrogate proxy function $ Q_{\eta} $, which simulates purification methods $P$ via a dynamic strength hyperparameter $\eta$. This surrogate proxy function includes a pair of projector and inverse projector that transform samples between the sample space and latent space. Purification is approximated through latent space filtering, which removes certain components of the latent representations, with the filtering strength controlled by a dynamically sampled hyperparameter. This proxy learning optimizes the following objective, given a classifier $f$, a training dataset $\mathcal{D}$, and a loss function $\mathcal{L}$:
\begin{equation}
\min_{f} \frac{1}{|\mathcal{D}|} \sum_{(x, y) \in \mathcal{D}} \mathcal{L}(f(Q_{\eta}(x), y)) .
\end{equation}

\subsubsection{Fourier Low-Pass Transform}
In practice, the choice of the surrogate proxy function can be diverse and one example is the low-pass Fourier filter~\cite{Brigham1988}, which removes high-frequency noises while preserving low-frequency features, approximating many purification mechanisms. In this case, the function consists of a discrete Fourier transform $\mathcal{F}$, an inverse discrete Fourier transform $\mathcal{F}^{-1}$ and a low-pass filter $G_R$, where $R$ represents the radius hyperparameter for controlling the filtering strength. The Fourier transform $\mathcal{F}$ projects an image $I(x, y)$ to a frequency-domain representation $F(u, v)$:
\begin{equation}
\mathcal{F} \left( I(x, y) \right)
        = \sum_{x=0}^{M-1} \sum_{y=0}^{N-1} I(x, y) e^{-i 2\pi \left( \frac{ux}{M} + \frac{vy}{N} \right)} ,
\end{equation}
whereas its inverse counterpart $\mathcal{F}^{-1}$ projects the samples from frequency space back to the image space:
\begin{equation}
\mathcal{F}^{-1}(F(u, v))
        = \frac{1}{MN} \sum_{u=0}^{M-1} \sum_{v=0}^{N-1} F(u, v) e^{i 2\pi \left( \frac{ux}{M} + \frac{vy}{N} \right)} .
\end{equation}
The low-pass frequency filter is constructed by
\begin{equation}
L_R (u, v) = 
\begin{cases} 
1 & \text{if } \sqrt{u^2 + v^2} \leq R , \\ 
0 & \text{if } \sqrt{u^2 + v^2} > R .
\end{cases}
\end{equation}
The transformed samples are then generated by 
\begin{equation}
I'(x, y) = \mathcal{F}^{-1} \left( L_R \left( \mathcal{F}( I(x, y) ) \right) \right) .
\end{equation}

\subsubsection{Gaussian Dynamic Sampling with Rejection}
To enable smooth domain adaptation, we further employ a probabilistic sampling technique for incrementally increasing the diversity of filtering strength as curriculum learning. We employ a Gaussian probability density function for sampling the radius hyperparameter:
\begin{equation}
\mathcal{G}(R) = \frac{1}{\sigma \sqrt{2\pi}} e^{-\frac{(R - \mu)^2}{2\sigma^2}}
\end{equation}
where the mean $\mu$ is set to the upper bound $R_{\text{max}}$, and the standard deviation $\sigma$ to the discrepancy between the upper and lower bounds $\| R_{\text{max}} - R_{\text{min}} \|$. To adjust the value within the range interval $[R_{\text{min}}, R_{\text{max}}]$, we incorporate a rejection sampling technique that resamples $R$ if it does not satisfy $ R_{\text{min}} \leq R \leq R_{\text{max}} $. The value of $ R_{\text{min}} $ decreases monotonically to simulate progressively stronger purification.

\section{Experiment}

We validated our method on the CIFAR-10 dataset~\cite{Krizhevsky2009} using the PreActResNet18 architecture~\cite{He2016}. From the training dataset, 1,000 images were used to generate real watermark evidence, while the remaining 49,000 served as standard training data. A Fourier-based low-pass filter was employed as the surrogate function for proxy learning, while this mechanism was omitted in non-proxy learning.
During training, the filter radius upper bound \(R_{\text{max}}\) was fixed at 25, while the lower bound \(R_{\text{min}}\) decreased by 2 every 60 epochs, starting at 21 and reaching 9 by the end of the 420 epochs. Non-proxy training underwent a similar number of epochs, converging to watermark embedding accuracy and model inference accuracy comparable to those of proxy training, with the training setup aligned to ensure controlled experimental conditions.
Assuming ideal conditions, attackers were granted sufficient time and resources to break 1,000 images, while the defense relied on a random oracle equivalent hash function~\cite{Bellare1993}. DiffPure, a diffusion-based purification method, was applied to counter forgeries by preventing fake watermarks from retaining falsified categories and also defend against adversarial examples~\cite{Nie2022}.
We generated 1,000 fake watermark evidence images and 10,000 adversarial test samples using targeted and untargeted \(L_2\) and \(L_{\infty}\) PGD attacks, respectively, matching the number of real watermarks and clean test samples. Perturbation budgets of 0.5 and 8/255 were chosen to balance impact and visual integrity. 
Watermark embedding and anti-forgery effectiveness were evaluated using verification accuracy, as shown in Table~\ref{tab:watermark_verification_accuracy}, defined as the proportion of 1,000 true/forged watermarks correctly verified as genuine. Similarly, model prediction and adversarial defense performance were assessed using inference accuracy, as shown in Table~\ref{tab:model_inference_accuracy}, defined as the proportion of 10,000 clean/adversarial samples correctly classified with true labels.
The visual effects of adversarial attacks and purification defenses on image appearance are illustrated in Figure~\ref{fig:image_demo}.

\begin{table}[h]
\centering

\begin{subtable}[t]{\linewidth}
\centering
\begin{tabular}{|>{\centering\arraybackslash}p{1.5cm}|>{\raggedright\arraybackslash}p{1.5cm}|>{\centering\arraybackslash}p{1.2cm}|>{\centering\arraybackslash}p{1.2cm}|}
\hline
\makecell{\textbf{Watermark} \\ \textbf{Evidence}} & \makecell{\textbf{Defense/} \\ \textbf{Pre-Process}} & \multicolumn{2}{c|}{\makecell{\textbf{Proxy} \\ \textbf{Learning}}}  \\ \cline{3-4} 
 &  & \textbf{With} & \textbf{Without} \\ \hline
\multirow{2}{*}{True} & None & 0.99 & 0.99 \\ \cline{2-4} 
 & Purified & 0.83 & 0.59 \\ \hline
\multirow{2}{*}{\makecell{Forged, \\ $L_{\infty}$}} & None & 1.00 & 1.00 \\ \cline{2-4} 
 & Purified & 0.12 & 0.12 \\ \hline
\multirow{2}{*}{\makecell{Forged, \\ $L_{2}$}} & None & 0.99 & 1.00 \\ \cline{2-4} 
 & Purified & 0.07 & 0.07 \\ \hline
\end{tabular}
\caption{Watermark Verification Accuracy}
\label{tab:watermark_verification_accuracy}

\end{subtable}

\vspace{0.3cm} 

\begin{subtable}[t]{\linewidth}
\centering
\begin{tabular}{|>{\centering\arraybackslash}p{1.5cm}|>{\raggedright\arraybackslash}p{1.5cm}|>{\centering\arraybackslash}p{1.2cm}|>{\centering\arraybackslash}p{1.2cm}|}
\hline
\makecell{\textbf{Test} \\ \textbf{Sample}} & \makecell{\textbf{Defense/} \\ \textbf{Pre-Process}} & \multicolumn{2}{c|}{\makecell{\textbf{Proxy} \\ \textbf{Learning}}} \\ \cline{3-4} 
 &  & \textbf{With} & \textbf{Without} \\ \hline
\multirow{2}{*}{Clean} & None & 0.94 & 0.94 \\ \cline{2-4} 
 & Purified & 0.90 & 0.87 \\ \hline
\multirow{2}{*}{\makecell{Adversarial, \\ $L_{\infty}$}} & None & 0.00 & 0.00 \\ \cline{2-4} 
 & Purified & 0.54 & 0.49 \\ \hline
\multirow{2}{*}{\makecell{Adversarial, \\ $L_{2}$}} & None & 0.01 & 0.00 \\ \cline{2-4} 
 & Purified & 0.73 & 0.70 \\ \hline
\end{tabular}
\caption{Model Inference Accuracy} 
\label{tab:model_inference_accuracy}
\end{subtable}

\caption{Evaluation of purification-agnostic proxy learning.}
\label{tab:experimental_result}
\end{table}

\begin{itemize}
    \item \textbf{Reliability of Watermark Verification}: Both proxy-trained and non-proxy-trained models achieved 94\% inference accuracy on clean test samples and 99\% verification accuracy on true watermark evidence, demonstrating effective watermark embedding while maintaining acceptable inference capability.

    \item \textbf{Vulnerability to Evidence Forgery}: Adversarial forgery samples achieved nearly 100\% watermark verification accuracy on undefended models, indicating that adversarial forgeries effectively compromised undefended black-box watermark mechanisms.
    
    \item \textbf{Effectiveness of Adversarial Purification}: The watermark verification accuracy of purified adversarial forgery samples dropped from nearly 100\% to 12\% for \(L_{\infty}\) and 7\% for \(L_2\), demonstrating the robustness of the purification framework against adversarial forgeries.

    \item \textbf{Robustness via Proxy Learning}: Compared to the non-proxy-trained model (59\% and 87\%), the proxy-trained model performed better after purification, achieving 83\% in true watermark verification and 90\% in clean sample inference, demonstrating strong resilience to the side effect of purification.
\end{itemize}

Additionally, the proxy-trained model showed higher inference accuracy than the non-proxy-trained model on purified \(L_2\) and \(L_{\infty}\) adversarial test samples, demonstrating the enhanced effectiveness of purification for the proxy-trained model.

\begin{figure}[htbp]
    \centering
    \includegraphics[width=0.75\linewidth]{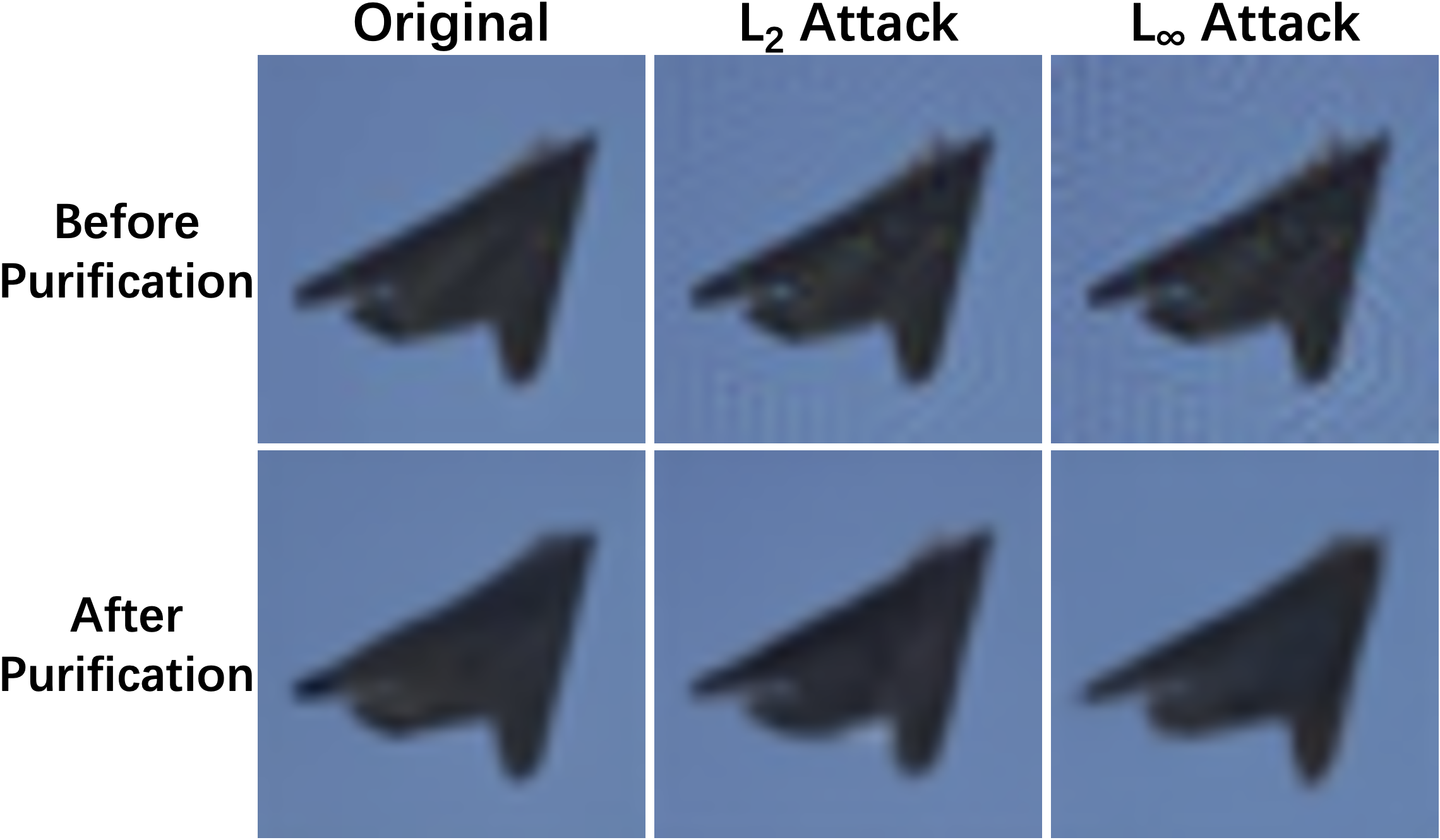}
    \caption{Demonstration of adversarial attack and defense.}
    \label{fig:image_demo}
\end{figure}

\section{Conclusion}

This work makes several significant contributions to the field of model watermarking, including the development of a self-authenticating black-box watermarking protocol utilizing hash techniques, an analysis of evidence forgery attacks and perturbation-based countermeasures and the proposal of purification-agnostic curriculum proxy learning for enhancing the reliability of watermark verification and overall model performance. Experimental results underscore the effectiveness of these approaches, highlighting their potential to significantly improve the performance of watermarked models. For future research, further exploration into various types of attacks and defenses are crucial for benchmarking watermarking security. Additionally, expanding the applicability of watermarking techniques to a broader range of AI models and exploring their robustness in more diverse operational environments would also be valuable directions for ongoing research.

\newpage
\balance

\end{CJK}
\end{document}